%% file: main.tex
\documentclass[journal]{IEEEtran}
\usepackage{algorithmic}
\usepackage{algorithm}
\usepackage{array}
\usepackage[caption=false,font=normalsize,labelfont=sf,textfont=sf]{subfig}
\usepackage{textcomp}
\usepackage{stfloats}
\usepackage{url}
\usepackage{cite}
\newcommand{\myparagraph}[1]{\vspace{0.1em}\noindent\textbf{#1}}
\usepackage[colorlinks,hyperindex,breaklinks]{hyperref}
\newcommand{\ie}{\textit{i}.\textit{e}.}
\newcommand{\eg}{\textit{e}.\textit{g}.}

\usepackage{pifont}
\newcommand{\cmark}{\ding{51}}%
\usepackage{graphicx}
\usepackage{colortbl}
\usepackage{arydshln}
\usepackage{multirow}
\usepackage{multicol}
\usepackage{mathrsfs}
\usepackage{bm}
\usepackage{amsfonts}
\usepackage{amssymb} 
\usepackage{ragged2e}
\newcommand\etc{etc\@ifnextchar.{}{.\@}}
\usepackage{amsmath}
\usepackage{amsthm}
\usepackage{verbatim}
\usepackage{latexsym}
\usepackage{silence}
\usepackage{booktabs}
\usepackage{makecell}
\def\BibTeX{{\rm B\kern-.05em{\sc i\kern-.025em b}\kern-.08em
    T\kern-.1667em\lower.7ex\hbox{E}\kern-.125emX}}
\usepackage{balance}
\begin{document}
\title{Boosting Weakly-Supervised Image Segmentation via Representation, Transform, and Compensator}
\author{Chunyan~Wang,
        Dong~Zhang,~\IEEEmembership{Member,~IEEE},
        Rui~Yan
\thanks{C. Wang is with the Department of Computer Science and Engineering, Nanjing University of Science and Technology, Nanjing 210094, China. E-mail:~carrie\_yan@njust.edu.cn. R. Yan is with the Department of Computer Science and Technology, Nanjing University, Nanjing 210023, China. E-mail:~ruiyan@nju.edu.cn. D. Zhang is with Department of Computer Science and Engineering, The Hong Kong University of Science and Technology, Hong Kong, China. E-mail:~dongz@ust.hk.}
\thanks{Corresponding author: Dong~Zhang and Rui~Yan.}
\thanks{The source codes have been released at:~\href{https://github.com/ChunyanWang1/RTC}{RTC}.}
}
\maketitle
\input{sections/0_abs}
\IEEEpeerreviewmaketitle
\input{sections/1_intro}
\input{sections/2_relatedwork}
\input{sections/3_methods}

\input{sections/4_experiments}

\input{sections/5_conclusion}
\bibliographystyle{IEEEtran}
\bibliography{IEEE_ref}
\end{document}

%% file: sections/0_abs.tex
\begin{abstract}
Weakly-supervised image segmentation (WSIS) is a critical task in computer vision that relies on image-level class labels. Multi-stage training procedures have been widely used in existing WSIS approaches to obtain high-quality pseudo-masks as ground-truth, resulting in significant progress. However, single-stage WSIS methods have recently gained attention due to their potential for simplifying training procedures, despite often suffering from low-quality pseudo-masks that limit their practical applications. To address this issue, we propose a novel single-stage WSIS method that utilizes a siamese network with contrastive learning to improve the quality of class activation maps (CAMs) and achieve a self-refinement process. Our approach employs a cross-representation refinement method that expands reliable object regions by utilizing different feature representations from the backbone. Additionally, we introduce a cross-transform regularization module that learns robust class prototypes for contrastive learning and captures global context information to feed back rough CAMs, thereby improving the quality of CAMs. Our final high-quality CAMs are used as pseudo-masks to supervise the segmentation result. Experimental results on the PASCAL VOC 2012 dataset demonstrate that our method significantly outperforms other state-of-the-art methods, achieving 67.2\% and 68.76\% mIoU on PASCAL VOC 2012 val set and test set, respectively. Furthermore, our method has been extended to weakly supervised object localization task, and experimental results demonstrate that our method continues to achieve very competitive results.
\end{abstract}

%% file: sections/1_intro.tex
\section{Introduction}
\label{intro}
\IEEEPARstart{S}{emantic} segmentation is a fundamental task in computer vision that involves assigning a class label to each pixel in an input image~\cite{noh2015learning,long2015fully}. Benefiting from the development of fully convolutional network, significant progress has been made in fully-supervised semantic segmentation methods~\cite{zhang2018context,zhong2020squeeze}. However, these approaches require extensive and meticulous pixel-level annotations, which can be expensive and time-consuming. To address this problem, an alternative approach called weakly-supervised image segmentation (WSIS) has been proposed~\cite{minaee2021image}. WSIS learns from less expensive and less elaborate supervisions, such as image-level class labels~\cite{ahn2018learning,xu2022multi,zhou2022regional}, bounding boxes~\cite{lee2021bbam,kulharia2020box2seg}, points~\cite{bearman2016s,qian2019weakly}, and scribbles~\cite{luo2022scribble,liang2022tree}. Among these, using image-level class labels as supervision is the most challenging for WSIS~\cite{wang2021exploring}. Our method focuses on this approach and aims to improve the performance of WSIS using image-level class labels.

\input{figs_tables/fig1.tex}
In the realm of WSIS, multi-stage frameworks have been widely used to achieve segmentation results. These frameworks typically involve training a multi-label classification model to generate classification activation maps (CAMs) that provide coarse object location cues. Pixel affinity networks or extra refinement procedures are then used to obtain more precise object regions as pseudo-masks. Finally, a fully supervised semantic segmentation model is trained with the pseudo-masks to achieve the segmentation results~\cite{zhou2016learning, ahn2018learning, ahn2019weakly, lee2021anti}. While multi-stage frameworks have been shown to yield significant performance improvements, they require designing multiple training networks, which can be time-consuming and sacrifice computation efficiency~\cite{araslanov2020single, zhang2020reliability}. To address this issue, single-stage WSIS methods have been proposed to jointly deal with image-level classification and pixel-level segmentation problems by sharing a backbone network~\cite{zhang2021adaptive, ru2022learning, bircanoglu2022isim, pan2022learning,ru2023token}. However, the segmentation performance of single-stage methods is inferior to that of multi-stage ones because it makes the supervision gap between classification and segmentation tasks more manifest. Recent studies have introduced matrix factorization, post-refinement strategies, image reconstruction or contrastive learning to improve the quality of CAMs and mitigate the supervision gap~\cite{du2022weakly, xie2022contrastive, zhou2022regional, xie2022clims, chen2022self,kweon2023weakly}. While multi-stage WSIS methods achieve higher segmentation performance than single-stage ones, they sacrifice training efficiency. On the other hand, some single-stage methods, such as RRM and SLRNet, as shown in Figure~\ref{fig1}, obtain good segmentation performance while bringing too high computational complexity, even more than the multi-stage ones~\cite{zhang2020reliability, pan2022learning}.

\input{figs_tables/fig2.tex}
Contrastive learning has been introduced into muti-stage weakly-supervised semantic segmentation methods, which can improve the feature representation while introducing relatively less computational complexity~\cite{du2022weakly,xie2022contrastive,zhou2022regional,xie2022clims}. Motivated by this, in this paper, we introduce contrastive learning into the single-stage weakly-supervised setting to provide pixel-level guidance information for learning accurate pseudo-masks online. However, two problems are encountered when introducing the contrastive learning mechanism: 1) under-activated or over-activated regions lead to unreliable class prototype estimation and inaccurate contrasts, as shown in the right side of Figure~\ref{fig2}, resulting in performance degeneration; and 2) the quality of pseudo-masks can deteriorate with increased training iterations, as shown in the left side of Figure~\ref{fig2}. To address these problems, we propose a single-stage WSIS method called boosting weakly-supervised image segmentation via representation, transform, and compensator (\emph{RTC}). \emph{RTC} comprises three fundamental modules: Cross-Representation Refinement (\emph{CRR}), Cross-Transform Regularization (\emph{CTR}), and Compensatory Loss (\emph{ComLoss}). \emph{CRR} expands the CAMs by leveraging different representations in the backbone layers to mine more reliable object regions. \emph{CTR} generates robust class prototypes by leveraging the semantic consistency between different representations from different affine transformations on the same image, which can alleviate the impact of over-activation on contrastive learning as well as identify more low-confident object regions that are hard to discriminate. \emph{ComLoss} refines the original CAMs and enhances the semantic representation by using the enhanced CAMs optimized by the global affinities of inter-images to feed back the rough CAMs. The combination of these modules enables us to obtain more accurate and complete CAMs in the classification branches, leading to high-quality pseudo-masks that can supervise the segmentation results online. As shown in Figure~\ref{fig2}, we can intuitively observe our method can alleviate the aforementioned problems and achieve significant segmentation performance. To demonstrate the generality of our proposed method, we also extend \emph{RTC} to weakly supervised object localization (WSOL). Experimental results demonstrate that our approach achieves satisfactory performance and has strong competitiveness compared to state-of-the-art models.

Besides, \emph{RTC} achieves comparable performance with most multi-stage and single-stage WSIS methods on PASCAL VOC 2012~\cite{everingham2010pascal}. Figure~\ref{fig1} shows the performance and complexity comparison of some advanced methods and our method on the PASCAL VOC 2012, which intuitively highlights the strength of our model. Our method can produce decent results with relatively less computational complexity. The main contributions of our work are: 
1) leveraging the \emph{CRR}, \emph{CTR} modules to alleviate the supervision gap between classification and segmentation tasks; and 2) proposing the \emph{ComLoss} module to further improve the quality of CAMs to prevent the pseudo-masks from becoming inferior with increased training iterations. The experimental results demonstrate that our proposed method achieves high qualities of CAMs on PASCAL VOC 2012~\cite{everingham2010pascal} for WSIS and CUB-200-2011~\cite{welinder2010caltech} for WSOL.

%% file: figs_tables/fig1.tex
\begin{figure}[tb]
\includegraphics[width=.48 \textwidth]{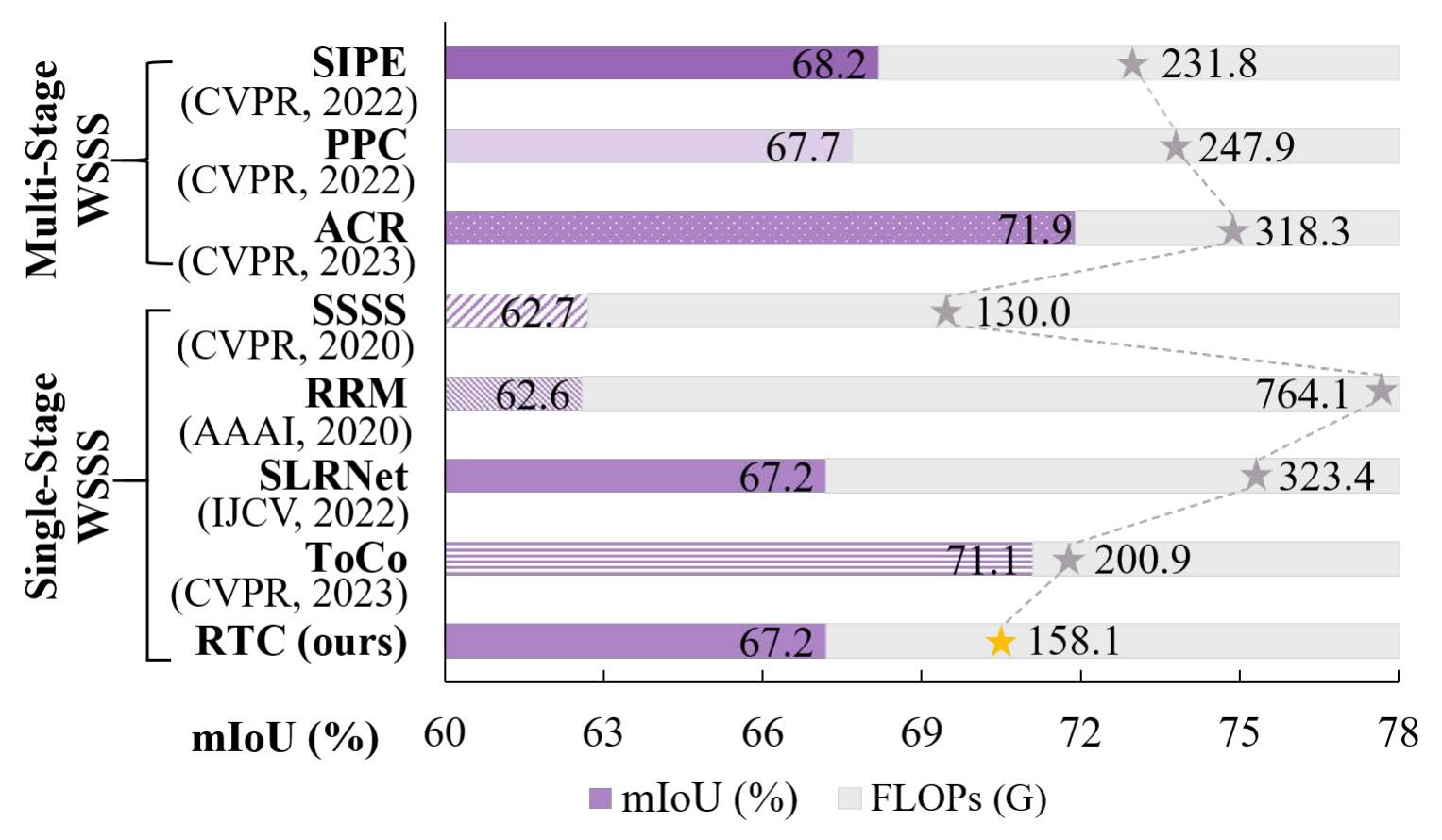}
\vspace{-4mm}
\caption{The segmentation accuracy and computational complexity of current weakly-supervised image segmentation (WSIS) methods on PASCAL VOC 2012~\cite{everingham2010pascal}. Multi-stage methods have demonstrated significant performance improvements, albeit at the cost of training efficiency. In comparison to other single-stage WSIS methods, our proposed approach achieves superior segmentation accuracy while exhibiting lower computational complexity. The computational complexity is measured in terms of FLOPs, and the input size is fixed at~$224\times 224$. }
\label{fig1}
\vspace{-2mm}
\end{figure}

%% file: figs_tables/fig2.tex
\begin{figure*}[tb]
\includegraphics[width=.99 \textwidth]{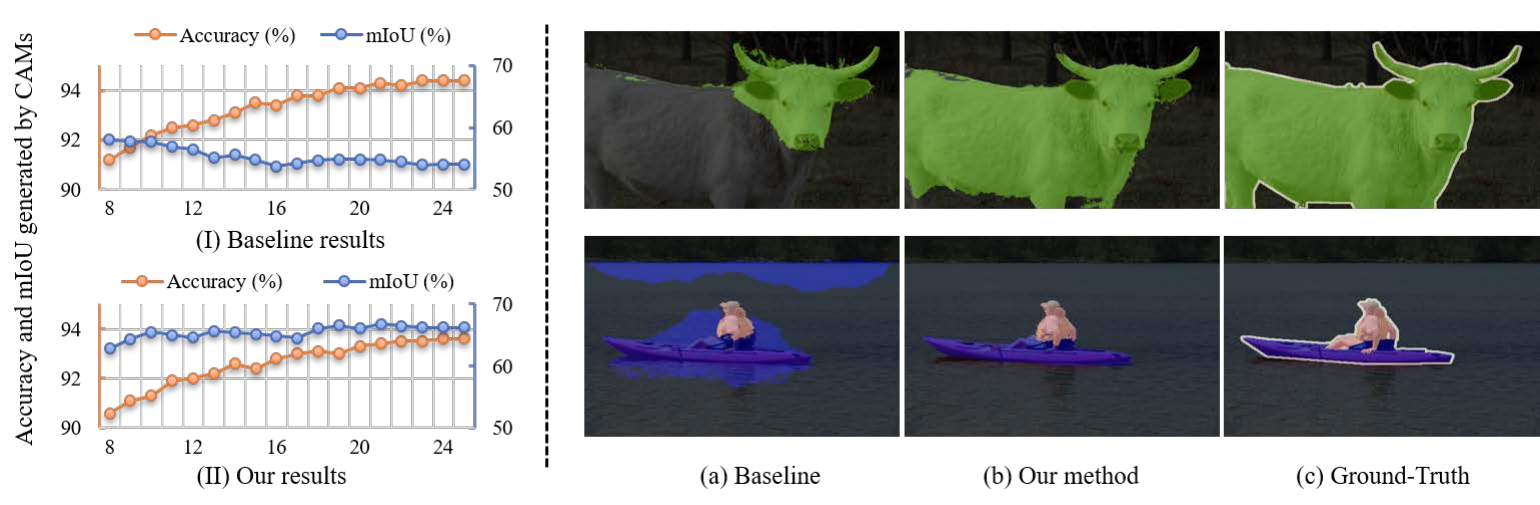}
\vspace{-4mm}
\caption{The primary motivation of this study is to address two key issues observed in the generated CAMs produced by the baseline method~\cite{du2022weakly}: (1) the presence of under-activation and over-activation (as shown in (a)), and (2) the degradation of CAMs quality with increasing iterations leading to inferior segmentation performance (as shown in (I)). In contrast to the baseline approach, our proposed method generates more complete and precise CAMs as pseudo-masks while maintaining high quality with increasing training iterations. It should be noted that the segmentation results of the baseline method presented here are based on our re-implementation of the method, which incorporates an additional segmentation branch in the PPC~\cite{du2022weakly} to achieve end-to-end weakly-supervised image segmentation. Our proposed method, referred to as \textbf{RTC}, outperforms the baseline method in terms of CAMs quality and segmentation accuracy.}
\label{fig2}
\vspace{-2mm}
\end{figure*}

%% file: sections/2_relatedwork.tex
\section{Related Work}
\subsection{Weakly-Supervised Semantic Segmentation (WSSS)}
To reduce the burden of collecting pixel-level annotations in fully-supervised semantic segmentation, various WSSS approaches have been proposed~\cite{zhang2021self,zhang2020causal,zhang2022deep}. For image-level WSSS, CAMs are commonly utilized to act as the original pseudo-masks. However, CAMs can only highlight the most discriminative regions of objects, hence most WSIS works \cite{zhang2020causal,lee2021reducing,chen2022self} utilize a three-stage learning model: initial CAMs generation, pseudo-masks refinement and fully-supervised semantic segmentation model training. Several methods rely on strategies like region erasing \cite{hou2018self,zhang2022unabridged,wei2017object} or growing \cite{huang2018weakly} to complement incomplete CAMs. Other methods focus on refining the CAMs by methods such as random walk \cite{bertasius2017convolutional} and pixel-level affinities \cite{ahn2018learning,zhang2023augmented}, to get precise pseudo-masks. However, due to the complicated training procedures of multi-stage training, some end-to-end models for WSIS have been proposed. Araslanov et al. \cite{araslanov2020single} propose the nGWP pooling and stochastic gate to cover more reliable regions of CAMs and learn the pixel affinities to refine them in a self-supervised way. Zhang et al. \cite{zhang2020reliability} decompose this WSIS problem into parallel classification and segmentation tasks, learning reliable pixel-level pseudo-masks in the classification branch to supervise the segmentation branch. Ru et al. \cite{ru2022learning} learn semantic affinity from pseudo-masks to supervise affinity learning in transformers in an end-to-end way. Chen et al. \cite{chen2021end} explore object boundaries explicitly to improve the recognition capability of a single model. Kho et al. \cite{kho2022exploiting} exploit shape cues to assist texture-based CNN, which provides the boundary information implicitly for generating precise pseudo-masks~\cite{zhang2022graph}. Unlike previous work, we introduce contrastive learning to provide additional supervision for improving the quality of CAMs online.

\subsection{Contrastive Learning in Semantic Segmentation }
Contrastive learning has gained popularity due to its significant progress in unsupervised/self-supervised tasks \cite{he2020momentum,khosla2020supervised,chen2022self}, which aims to minimize the distance of samples from the positive (similar) pair while maximizing the distance from the negative (different) pair. Recently, several works \cite{wang2021exploring,zhao2022cross,du2022weakly} have proposed to perform contrastive learning in semantic segmentation, achieving promising performance in fully-supervised \cite{wang2021exploring,zhang2022deep}, semi-supervised \cite{zhao2022cross}, and weakly-supervised \cite{du2022weakly} settings. Wang et al. \cite{wang2021exploring} propose pixel-level contrast learning to address inter-class dispersion and intra-class compactness, boosting the segmentation performance. Pissas et al. \cite{pissas2022multi} introduce multi-scale and cross-scale contrastive learning on the encoder's features to enhance the discriminative ability of the model. Zhao et al. \cite{zhao2022cross} employ cross-scale contrastive learning to explore the similarity cues between global and local features, which effectively learn the local characteristics. Xie et al. \cite{xie2022c2am} learn class-agnostic activation maps by foreground-background contrastive loss to discover more object regions. Du et al. \cite{du2022weakly} adopt the pixel-to-prototype contrast providing additional pixel-level supervision for enhancing the qualities of CAMs. Inspired by \cite{du2022weakly}, we introduce the pixel-to-prototype contrastive learning architecture to improve the quality of CAMs online as well as generate less computational complexity.  While different from \cite{du2022weakly}, we leverage the other view's projected features to estimate the class prototypes of the current view, which can enhance the robust feature representation of class prototypes meanwhile avoid the complicated sampling strategies. 

\subsection{Consistency Regularization}
Consistency regularization has been extensively studied in a range of label-scarce segmentation tasks, such as semi-supervised learning \cite{ke2020guided,ouali2020semi}, weakly supervised learning \cite{wang2020self,chen2022self}, and unsupervised learning \cite{grill2020bootstrap}, which certifies the effectiveness of improving feature representation. It enforces the network to produce consistent predictions or intermediate feature outputs under different perturbations, such as image augmentation \cite{ke2020guided} and feature perturbations \cite{ouali2020semi}. For example, to address the consistent prediction of pyramid predictions, Luo et al. \cite{luo2022semi} learn to minimize the discrepancy between each scale prediction and their average. Wang et al. \cite{wang2020self} adopt consistency regularization on CAMs generated from two transformed input images to provide self-supervision learning. Chen et al. \cite{chen2022self} propose a consistency loss to optimize and self-correct the feature representation. Inspired by the work of Wang et al. \cite{wang2020self}, we consider the distinctive activation regions of differently augmented inputs and introduce the cross-transform regularization module and compensatory loss to learn complete and precise object regions, which construct additional supervision information for model learning with only image-level class labels.

%% file: sections/3_methods.tex
\section{Methodology}

\input{figs_tables/fig3}
\subsection{Preliminaries}
\label{sec3:1}
Contrastive learning~\cite{wang2021exploring,zhao2022cross,du2022weakly} aims to pull pixels of the same class together and push away pixels of different classes in the projected feature space for semantic segmentation. We adopt the pixel-to-prototype contrastive loss proposed in PPC~\cite{du2022weakly} to learn more complete object regions, which can be formulated as follows:
\begin{gather}
{\bm L}_{contrast}({\bm X}_{i},{\bm y}_{i},{\bm P})=-log \frac{exp({\bm X}_{i}\cdot{\bm P}_{{\bm y}_{i}}/ \tau)}{\sum_{{\bm p}_c\in {\bm P}} exp({\bm X}_{i}\cdot{\bm p}_c / \tau) }  \label{1} \\
{\bm P}_{c}=\frac{\sum_{j\in \Phi_c }{\bm M}_{c,j}{\bm X}_{j}}{\sum_{j\in \Phi_c}{\bm M}_{c,j}} \label{2} 
\end{gather}
Here~${\bm X}_{i}$ is the feature representation of any pixel~$i$ in a feature map.~${\bm M}$ represents CAMs and~$y=argmax(\bm M)$.~${\bm P_{c} \in \mathbb{R}^{1\times D}}$ represents a prototype of class~$c$~($\in \left [ 1,\cdots, C \right ]$,~$C$ is the set of classes in the dataset), which is a $D$-dimension vector. $\bm P$ is the collection of all class prototypes.~${\bm P}_{{\bm y}_{i}}$ means that the pseudo label of each pixel~$i$ is represented by the corresponding class prototypes.~$\Phi_c$ is the collection of top~$K$ pixels of class~$c$ with high scores in the~${\bm M}_{c}$.~$K$ is a hyperparameter.~$\tau$ is a temperature parameter. From Eq. (\ref{1}) and Eq. (\ref{2}), we know the quality of CAMs influences the learning of class prototypes while the class prototypes play an important role in the effectiveness of contrastive learning. Thus, it is crucial to develop methods that can generate high-quality CAMs and robust class prototypes.

\subsection{Overall}
Figure~\ref{overview} shows an overall architecture of \textbf{RTC}, which can boost the online performance of pseudo-masks. First, an input image~${\bm I} \in \mathbb{R}^{3\times H\times W}$~(where~$H$ and~$W$ are the height and width of the image spatial size,~$3$ denotes the channel size) and a rescaled input image are sent into the share-weighted encoder network to extract features~${\bm F} \in \mathbb{R}^{d\times h\times w}$ (where~$d$ is the channel size and $h\times w$ denotes the spatial size), which are 1/8 spatial size of the input image, for generating CAMs and segmentation results respectively. In the classification branches, the extracted features are first utilized to produce projected features~${\bm X}^{1}$ and~${\bm X}^{2}$ and then generate the CAMs. Next, CRR modules are employed to expand the reliable object regions of CAMs by leveraging the different feature representation information of different layers. After that, we combine the other view's projected features with the self-view CAMs to learn the robust class prototypes by the CTR modules, which are utilized to alleviate the problem of over-activation and identify more low-confident object regions. To mitigate the inferior qualities of CAMs with the increased training iterations, ComLoss is proposed to exploit the global semantic affinities between a part of projected features corresponding to highly confident class regions and the whole projected features of the other view in order to enhance the original CAMs. Finally, we adopt the post-processing method~\cite{xu2023self} on refined CAMs to achieve the pseudo-masks for supervising the segmentation results.

\subsection{Cross-Representation Refinement (CRR)}
\label{sec3:2}
For a given image, the feature maps~${\bm F}$ are firstly extracted by a pre-trained classification network. Then a $1\times 1$ convolution layer followed by a ReLU function is adopted to generate the projected features~${\bm X} \in \mathbb{R}^{256\times h\times w}$ for the following CAMs generation and class prototypes exploration. After that, the features~${\bm X}$ are forwarded to a global average pooling followed by a fully connected layer whose weights are represented as~${\bm w}_{\theta} \in \mathbb{R}^{C\times 256}$ for classification. Thus, CAMs~${\bm {\tilde{M}}} \in \mathbb{R}^{C\times h\times w}$ are obtained by :
 \begin{gather}
 {\bm X}={Conv}({\bm F} )\label{3} \\
 {\bm {\tilde{M}}}=ReLU({\bm w}_{\theta}{\bm X}) \label{4}
 \end{gather}
where~$Conv$ means a~$1\times 1$ convolution layer.~${\bm w}_{\theta}$ is the weight of the last fully connected layer of the network.
 
However, CAMs only highlight the discriminative regions of objects. In order to mine more highly confident object regions, we resort to leveraging the semantic and detailed information embodied in deep and shallow features from different layers of a backbone to complement the incomplete CAMs as shown in Figure~\ref{overview}.
 Concretely, we concatenate the input~${\bm I}$ and features~${\bm F}_{4}$ from layer $4$ followed by a $1\times 1$ convolution layer to reduce the dimension and produce the shallow features~${\bm F}_{\mathrm s}\in \mathbb{R}^{d_1\times h\times w}$. Similarly, the features~${\bm F}_{5}$ and~${\bm F}_{6}$ from layer $5$ and layer $6$ are concatenated as deep features~${\bm F}_{\mathrm d}\in \mathbb{R}^{d_2\times h\times w}$.  
\begin{gather}
{\bm F}_{\mathrm s}={Conv}(Cat \left [{Conv}({\bm F}_{4}),I \right ]) \label{5} \\
{\bm F}_{\mathrm d}={Conv}(Cat \left [{Conv}({\bm F}_{5}),{\bm Conv}({\bm F}_{6}) \right ]) \label{6}
 \end{gather}
 Where~$Cat$ denotes the concatenation operation. After that, we calculate the affinities of pixels in shallow and deep features respectively, and then normalize them to $\left [ 0,1 \right ] $. The inter-pixel affinities $\bm A \in \mathbb{R}^{hw\times hw}$ can be calculated as follows:
 \begin{gather}
{\bm A}(i,j)=ReLU(\frac{{\bm F}^{T}(i){\bm F}(j)}{\left \| {\bm F}(i) \right \| \cdot \left \| {\bm F}(j) \right \|}) \label{7}
 \end{gather}
 where~${\bm F}$ represents the shallow or deep features, with spatial position index $i$ and $j$. A ReLU function is used to suppress the negative values. Finally, refined CAMs~${\bm M} \in \mathbb{R}^{C\times h\times w}$ are obtained by multiplying the original CAMs with the inter-pixel affinities of~${\bm A}_{s}$ and~${\bm A}_{d}$ respectively and adding together as follows:
 \begin{gather}
 {\bm M}=\frac{({\bm A}_{s} + {\bm A}_{d})}{2} {\bm {\tilde {M}}} \label{8}
\end{gather}

\subsection{Cross-Transform Regularization (CTR)}
\label{sec3:3}
After expanding the discriminative seed region by CRR, the under-activation problem can be alleviated. However, using solely CRR, the refined CAMs still face the problems of over-activation and low-confident regions that can not be identified, which affects the learning of class prototypes for contrastive learning. To address this issue, we propose the CTR module to learn robust class prototypes.

CAMs are sensitive to different spatial sizes of input images~\cite{wang2020self}, thus spatial augmentation operation~${\bm A(\cdot)}$ is first utilized to obtain another rescaled input as shown in Figure~\ref{overview}. Next, we can generate two projected features~${\bm X}^{1}$ and~${\bm X}^{2}$, refined CAMs~${\bm M}^{1}$ and~${\bm M}^{2}$. After that, we utilize the other view's projected features to interact with the self-view refined CAMs under the guidance of semantic consistency between~${\bm M}^{1}$ and~${\bm M}^{2}$ to learn robust class prototypes. Specifically, we resize and normalize the two projected features to the same size and interact with the cross-view refined CAMs to calculate each class prototypes~${\bm P}_{c}^{1}$ and~${\bm P}_{c}^{2}$. It can be formulated as follows:
\begin{gather}
{\bm P}_{c}^{1}=\frac{\sum_{i\in \Phi_c }{\bm M}^1_{c,i}{\bm X}^2_{i}}{\sum_{i\in \Phi_c }{\bm M}^1_{c,i}} \label{9} \\
{\bm P}_{c}^{2}=\frac{\sum_{i\in \Omega_c }{\bm M}^2_{c,i}{\bm X}^1_{i}}{\sum_{i\in \Omega_c }{\bm M}^2_{c,i}} \label{10}
\end{gather}
where~$\Phi_c$ and~$\Omega_c$ is the collection of top~$K$ pixels of class~$c$ in the~${\bm M}_{c}^1$ and~${\bm M}_{c}^2$, respectively. After obtaining two class prototype representations, we combine them with the contrastive loss shown in Eq. (\ref{1}) respectively to encourage the pixel-wise feature to be close to the corresponding class prototypes and far away from other class prototypes. Instead of using the self-view projected features to calculate with the self-view CAMs, cross-view ones can focus on the pixels hard to identify and be helpful to prevent the influence of noisy (over-activated) labels on contrastive learning (discussed in the Table~\ref{tab2} of the experiment). 

Besides, it can be helpful in learning more precise and complete object regions by applying the other view's pseudo-masks or prototypes in the current view in Eq. (\ref{1}) to force semantic consistency. Thus, we adopt the intra-view and cross-view contrast losses to improve the contrastive capability like~\cite{du2022weakly}.
Different from PPC~\cite{du2022weakly}, we leverage the other view's projected features combined with self-view CAMs to estimate the class prototypes, so as to reduce the influence of noise or weak (low-confident) labels and improve the robustness of feature representation of class prototypes meanwhile avoiding the complicated sampling strategies. 

\subsection{Compensatory Loss (ComLoss)}
\label{sec3:4}

The motivation of our ComLoss is to utilize the global semantic affinities to further refine the rough CAMs and then feed back the original CAMs~${\bm {\tilde {M}}}$, which can avoid the decreased qualities of CAMs caused by the increased training iterations. Specifically, to make full use of the projected features from two views, we learn global semantic relations from them to complement the original CAMs by calculating the affinities between a part of cross-view projected features corresponding to the highly confident object regions of the self-view CAMs and the whole cross-view projected features in a training batch. Concretely,  affinities~${\bm A}_{c}^1$ between the normalized whole projected features~${\bm X}_{c}^2$ and partial features~${\bm X}^2_{c,i}$ corresponding to the position of top~$K$ highly confident pixels in~${\bm M}_{c}^{1}$ in a training batch are firstly calculated to capture the global semantic relations. Then, the highest confident~${\bm M}_{c,i}^1$ multiplies with the affinities to capture other object regions that are similar to the highest confident object regions. After that, we add it with refined CAMs to obtain compensatory CAMs~${\bm {\hat{M}}_{c}^1}$. Finally, we complement the knowledge learned from one view to another through a~${\bm L}_1$ normalization loss between~${\bm {\hat{M}}}^1$ and~${\bm {\tilde{M}}^{1}}$.

\begin{gather}
    {\bm A}_{c}^1=\sigma (\frac{{\bm X}_{c}^2}{\left \| {\bm X}_{c}^2 \right \|}* (\frac{{\bm X}^2_{c,i}}{\left \| {\bm X}^2_{c,i}  \right \|})^T ),i \in \Phi_{c}  \label{11} \\
    {\bm {\hat{M}}}_{c}^1={\bm A}_{c}^1*{\bm M}^1_{c,i}+{\bm M}_{c}^1,i \in \Phi_{c}\label{12} \\
    {\bm L}_{com1}=\sum_{c\in C }\left | {\bm {\hat{M}}_{c}^1} -{\bm {\tilde {M}}_{c}^{1}} \right |  \label{13}
\end{gather}
Here~$\sigma$ represents the softmax function.~${\bm A}_{c}^1 \in \mathbb{R}^{1 \times nhw} $ is a global pixel affinities of class~$c$ in a training batch, where~$n$ is the batch size. Similarly, we can obtain the other view's global pixel affinities~${\bm A}_{c}^2$ and compensatory loss~${\bm L}_{com2}$. The total compensatory loss~$ {\bm L}_{com}={\bm L}_{com1}+{\bm L}_{com2}$.

\subsection{Overall Loss}\label{sec3:5}
We apply the post-processing refinement method proposed in~\cite{xu2023self} to refine the~${\bm M}^{1}$, which are utilized as the online pseudo-masks to supervise the segmentation result. The overall loss function ${\bm L}_{t}$ is:
\begin{gather}
    {\bm L}_{t}={\bm L}_{cls}+{\bm L}_{er}+{\bm L}_{ecr}+\alpha {\bm L}_{contrast}+\beta {\bm L}_{com}+{\bm L}_{seg}
\end{gather}
where~$\alpha$ and~$\beta$ are hyper-parameters to balance the losses in the experiment. ${\bm L}_{er}$ and ${\bm L}_{ecr}$ losses are the equivariant regularization loss and equivariant cross loss proposed in SEAM~\cite{wang2020self} to balance the input image and a rescaled images.

%% file: figs_tables/fig3.tex
\begin{figure*}[!t]
\centering
\includegraphics[width=.98\textwidth]{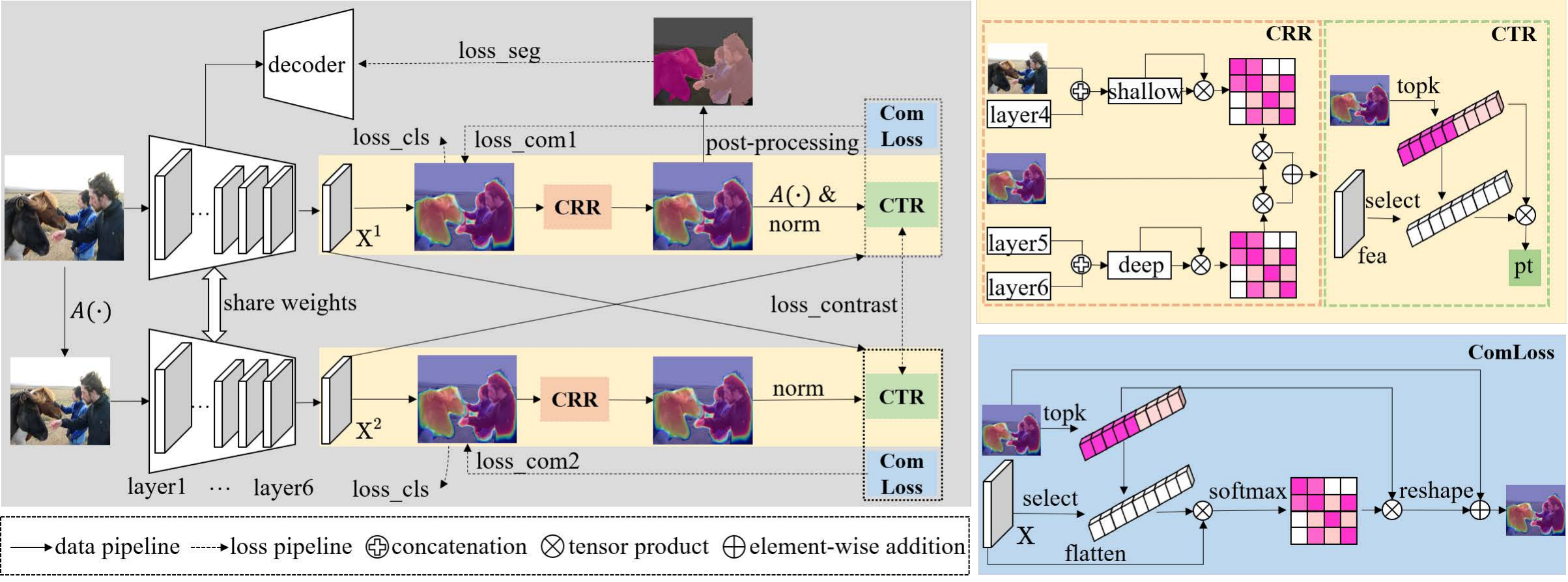}
\vspace{-2mm}
\caption{An overview of our proposed Representation, Transform, and Compensator (\textbf{RTC}), which involves a spatial transformation operation $A(\cdot)$ for image augmentation, as well as three progressive modules (CRR, CTR, and ComLoss) for improving the quality of CAMs and generating high-quality pseudo-masks for end-to-end supervision of the segmentation result. The CAMs quality improvement is achieved through the use of L2 normalization and the aforementioned modules. For concision, the pipeline of features from layer $4$, $5$, $6$, and the input image to the CRR module is omitted in the left image.} 
\label{overview}
\vspace{-2mm}
\end{figure*}

%% file: sections/4_experiments.tex
\section{Experiments}
\subsection{Dataset and Evaluation Metrics}\label{sec4:1}

\myparagraph{Datasets.}
We evaluate our method in the WSIS task and WSOL task based on PASCAL VOC 2012~\cite{everingham2010pascal} and CUB-200-2011~\cite{welinder2010caltech} datasets, respectively. PASCAL VOC 2012 has~$21$ classes (including one background) of objects in total of~$4,369$ images, which are split of~$1,464$ images for training,~$1,449$ images for validation and~$1,456$ images for testing, respectively. Following the common practice in semantic segmentation, the augmented annotations from SBD~\cite{hariharan2011semantic} are used for an experimental comparison that has~$10,582$ training images. CUB-200-2011~\cite{welinder2010caltech} contains~$200$ bird species, and it has~$5994$ images for training and~$5794$ for testing. 

\myparagraph{Evaluation Metrics.}
For WSIS, we use the mean Intersection-over-Union (mIoU) as the evaluation metrics to evaluate the segmentation results. Params (for parameters) and FLOPs (for computational complexity) are also provided for evaluating the efficiency of our method. For WSOL, we use Top-1 localization accuracy (Top-1 Loc), Top-5 localization accuracy (Top-5 Loc) and localization accuracy with ground-truth class (GT Loc) as our evaluation metrics. We also take the MaxBoxAccV2~\cite{choe2022evaluation} with multiple IoU thresholds $\delta \in \{0.3,0.5,0.7 \}$ to measure the localization accuracy without considering the classification result.

\subsection{Experimental Settings}
We adopt the commonly used ResNet38~\cite{wu2019wider} pre-trained on the ImageNet~\cite{deng2009imagenet} as the backbone with~$output\_stride=8$ for the WSIS task.
The images are cropped by~$448\times 448$ and~$128\times 128$ as two views' network inputs following the strategies of ~\cite{du2022weakly}. We extract the features from layer~$4$,~$5$,~$6$ and reduce their channel dimension to~$64$,~$128$,~$128$ respectively. We adopt~$256$ as the dimension of projected features for contrastive learning. Following~\cite{du2022weakly,kim2022bridging}, we set~$\alpha=0.1$ and~$\beta =1$. The initial learning rates are ~$ 0.01$. Our model is trained using the Poly optimizer with batch size~$8$, momentum of~$0.9$ and weight decay~$5e-4$, in an end-to-end manner for~$25$ epochs total. Concretely, we first train~$7$ epochs using all losses except segmentation loss and then switch on the segmentation loss for the remaining~$18$ epochs. For the WSOL task, we evaluate the effectiveness of our method based on the Bridging method~\cite{kim2022bridging} as an extension work, which utilizes the VGG16~\cite{simonyan2015very} and ResNet50~\cite{he2016deep} as the backbone.

\input{figs_tables/fig4}

\input{figs_tables/table1}

\subsection{Ablation Study}\label{sec4:2}
Our main ablation study is based on the WSIS task. The quantitative results of the ablation study are reported in Table~\ref{tab1} to investigate the effectiveness of each component as well as their combination in the qualities of CAMs and the final segmentation results on the \emph{train} and \emph{val} set of PASCAL VOC 2012. Figure~\ref{fig4} gives the visualization comparisons generated by the baseline and our \textbf{RTC} method on the \emph{train} and \emph{val} sets of PASCAL VOC 2012. We can obviously observe that our proposed modules can progressively learn more precise and complete object areas, which intuitively verifies the effectiveness of our method. We will analyze the effectiveness of each component in detail in the following section.

\input{figs_tables/table2}

\myparagraph{Effectiveness of CRR and CTR.}
As shown in Table~\ref{tab1}, we can observe that compared with the baseline, the model boosts the qualities of CAMs and improves by~$4.29\%$ and~$3.33\%$ mIoU with the help of CRR on the \emph{train} and \emph{val} set of PASCAL VOC 2012, respectively. Besides, with the improvement of qualities of CAMs, the model can achieve a segmentation performance by~$52.71\%$ mIoU on the \emph{val} set of PASCAL VOC 2012, which outperforms the baseline over~$5.06\%$ mIoU. When both CRR and CTR are implemented on the baseline model, the results of CAMs can achieve~$63.71\%$ and~$62.15\%$ mIoU on the \emph{train} and \emph{val} sets of PASCAL VOC 2012. Compared with the baseline, our model outperforms it over~$13.61\%$ mIoU. These results demonstrate CAMs can be expanded by utilizing the different feature representations from a backbone, further refined by learning robust class prototype representations for contrastive learning. It is noted that the segmentation result with the help of CTR and CRR achieves only~$57.09\%$ mIoU due to the qualities of CAMs being influenced by the increased training iterations, leading to inferior segmentation results.

\myparagraph{Effectiveness of ComLoss.}
In Table~\ref{tab1}, we can observe that with the help of ComLoss, the qualities of CAMs can achieve from~$63.71\%$ to~$65.50\%$ mIoU on the \emph{train} set of PASCAL VOC 2012. Most importantly, the segmentation performance can be improved from~$57.09\%$ to~$65.34\%$ mIoU. It demonstrates that ComLoss can not only alleviate the inferior qualities of CAMs with the increased training iterations but also enhance the feature representation of CAMs.

\input{figs_tables/table3}
\input{figs_tables/fig5}
\input{figs_tables/table4}

\subsection{Complexity Analysis}\label{complexity}
We conducted a complexity analysis of our method, as shown in Table~\ref{tab1}. Our CRR module introduces $0.73$ M parameters and $2.27$ G FLOPs into the model, while the CTR and ComLoss modules do not introduce any additional parameters or computational complexity. To further evaluate the efficiency of our method, we compared it with other state-of-the-art methods in Figure~\ref{fig1}. We observed that multi-stage WSIS methods such as SIPE~\cite{chen2022self}, PPC~\cite{du2022weakly}, and ACR~\cite{kweon2023weakly} achieve high segmentation performance but increase computational complexity. On the other hand, single-stage RRM~\cite{zhang2020reliability} and SLRNet~\cite{pan2022learning} methods also achieve good performance, but their computational complexities are even higher than that of multi-stage methods, reaching $764.1$ G and $323.4$ G FLOPs, respectively. In comparison, our proposed method achieves excellent segmentation performance while bringing less computational complexity. Although ToCo~\cite{ru2023token} achieves higher segmentation performance than our \textbf{RTC}, its computational complexity is $42.8$ G higher than our method. These results further validate the effectiveness and efficiency of our proposed modules in improving the quality of CAMs.

\subsection{Superiority of CRR and CTR}\label{sec4:3}
Our proposed CRR and CTR aim to cover more confident object regions by leveraging the different representations in the backbone layers and learning robust class prototypes from cross-view projected features, respectively. In this section, to further demonstrate the superiority of CRR and CTR modules, we compare them against other refinement strategies.

\myparagraph{Superiority of CRR.}
Our CRR utilizes the affinity matrices of deep and shallow feature representations from the backbone in parallel to refine the CAMs. In order to show the superiority of our CRR, we compare it with the PCM~\cite{wang2020self} (which first fuses the features from layer~$4$,~$5$ and images, then uses the fused features to refine the CAMs). As shown in Table~\ref{tab2}, we can observe that our combination of CRR and CTR can achieve a segmentation result of~$57.09\%$ mIoU on the \emph{val} set of PASCAL VOC 2012, while the combination of PCM and CTR can only achieve~$52.19\%$ mIoU. It validates the superiority of CRR in separately utilizing the different representations from the backbone to expand the CAMs rather than the fused deep and shallow features.

\myparagraph{Superiority of CTR.}
CTR employed the cross-view projected features and self-view CAMs to calculate the class prototypes, which are robust to the noise and weak labels. To validate the superiority of CTR, we compare it with the self-view transform regularization STR (which utilizes the self-view projected features and CAMs to estimate the class prototypes) method. From Table~\ref{tab2}, we can observe that the CTR can obtain both higher qualities of CAMs and segmentation results than the self-view transform regularization STR, which validates the superiority of adopting the cross-view projected features to estimate the class prototypes rather than the self-view ones. The reason is that the utilization of cross-view projected features can alleviate the impact of noisy labels on contrastive learning as well as give more focus on the less discriminative object regions, leading to the discovery of more accurate and complete object regions.

\subsection{Comparisons with State-of-the-art WSIS Methods}\label{sec4:4}
In this section, we make result comparisons between our \textbf{RTC} and other state-of-the-art WSIS methods quantitatively and qualitatively.

\myparagraph{Quantitative results.}
As shown in Table~\ref{tab3}, we show segmentation performance on the \emph{val} and \emph{test} sets of PASCAL VOC 2012~\cite{everingham2010pascal}. Our single-stage method achieves~$65.34\%$ and~$66.90\%$ mIoU on the VOC \emph{val} and \emph{test}\footnote{http://host.robots.ox.ac.uk:8080/anonymous/B9PIXX.html} sets. Compared to the AA\&LR~\cite{zhang2021adaptive} and SSSS~\cite{araslanov2020single} methods, our method can still outperform it by~$1.44\%$ and~$2.64\%$ mIoU respectively even without the help of the CRF operation. And with the help of the CRF, our \textbf{RTC} can achieve~$67.20\%$ and~$68.76\%$ mIoU on the \emph{val} and \emph{test}\footnote{http://host.robots.ox.ac.uk:8080/anonymous/HXCNME.html} sets of VOC 2012, which outperforms most of end-to-end methods. Specifically, our method outperforms AFA~\cite{ru2022learning} with MiT-B1 as a backbone, which improves by~$1.2\%$ and~$2.46\%$ mIoU on the \emph{val} and \emph{test} sets respectively. ToCo~\cite{ru2023token} method is superior to our \textbf{RTC} because it learns better global dependency by transformer as a backbone as well as low confident object regions by global-local contrastive learning, which utilized more computational complexity than our method.
Furthermore, our method achieves higher mIoU even than some multi-stage methods using saliency maps as extra supervision data, \eg, EPS~\cite{lee2021railroad} and OAA+~\cite{jiang2021online}. These results demonstrate the effectiveness of our method in covering more confident object regions.

We also extend our \textbf{RTC} to a two-step framework to show the superiority and scalability of our method. We first adopted our single-stage framework to produce the pseudo-masks for the training dataset. After that, we utilized the generated pseudo-masks to train the Deeplab~\cite{liang2015semantic} with ResNet38 and ResNet101 backbone respectively. For the ResNet38, the final results can achieve~$70.21\%$ and~$70.76\%$ mIoU on the \emph{val} and \emph{test}\footnote{http://host.robots.ox.ac.uk:8080/anonymous/YHPKJJ.html} sets of VOC 2012 respectively shown in Table~\ref{tab3}, which outperforms most of the multi-stage WSIS methods. In addition, the segmentation results can arrive at a higher performance with ResNet101, which can further obtain~$71.56\%$ and~$72.33\%$ mIoU on the \emph{val} and \emph{test}\footnote{http://host.robots.ox.ac.uk:8080/anonymous/NOVA6X.html} sets of VOC 2012. 
Compared to the SAM method~\cite{kirillov2023segany,jiang2023segment}, our method even outperforms it by $2.56\%$ and $3.63\%$ mIoU on the \emph{val} and \emph{test} sets of VOC 2012 supervised by the point-level labels. Noted that the results of SAM method here
are also obtained by a two-step framework: SAM~\cite{kirillov2023segany} first generates the pseudo-masks by the provided weak supervision information and then trains a fully supervised semantic segmentation network based on the Deeplab with ResNet101 as the backbone. These results further demonstrate our \textbf{RTC} can learn more accurate and complete object regions.

\myparagraph{Qualitative results.} The results of our proposed method and other state-of-the-art methods, including SSSS~\cite{araslanov2020single}, AFA~\cite{ru2022learning}, and SLRNet~\cite{pan2022learning}, on the \emph{val} set of PASCAL VOC 2012~\cite{everingham2010pascal} are presented in Figure~\ref{fig5}. Our method outperforms the other methods in terms of segmentation accuracy, when using the same backbone. All methods except AFA use the ResNet38~\cite{wu2019wider} as a backbone. Our \textbf{RTC} method still outperforms AFA, which uses the stronger MiT-B1 as the backbone, by a large margin on most classes, such as ``bicycle'', ``car'', and ``train''. This demonstrates that our method is effective in improving the quality of CAMs as online pseudo-masks for supervising the segmentation result.

\input{figs_tables/fig6}

\subsection{Extensions on Weakly Supervised Object Localization}
Our proposed method is a general approach that can boost the performance of models for all weakly supervised visual tasks. To evaluate the universality of our proposed method, we extend it to the weakly supervised object localization task. We conduct experiments on the CUB-200-2011 dataset, which reports the location performance by the MaxBoxAccV2~\cite{choe2022evaluation} scores and location accuracy, as presented in Table~\ref{tab4}. Our proposed \textbf{RTC} achieves MaxBoxAccV2 scores of $84.31\%$ and $81.90\%$ mIoU on the \emph{test} set based on the VGG16 and ResNet50 backbones, respectively. Notably, our method can improve the score with the IoU threshold of 0.7 by $9.20\%$ and $16.00\%$ with VGG16 and ResNet50, respectively, indicating that our method can accurately identify the less discriminative regions of target objects. Furthermore, our proposed method also achieves better location performance by considering the classification results compared to the baseline model, thereby validating that our method can accurately complement the incomplete object regions. The visualization results of the baseline method and our method on the \emph{test} set of CUB-200-2011~\cite{welinder2010caltech} are shown in Figure~\ref{fig6}. As can be seen, our method intuitively complements the discriminative object regions of CAMs and locates the object more accurately, which further demonstrates the effectiveness of our method. These results highlight the potential of our proposed approach in improving the performance of weakly supervised object localization.

%% file: figs_tables/fig4.tex
\begin{figure*}[tb]
\includegraphics[width=.98 \textwidth]{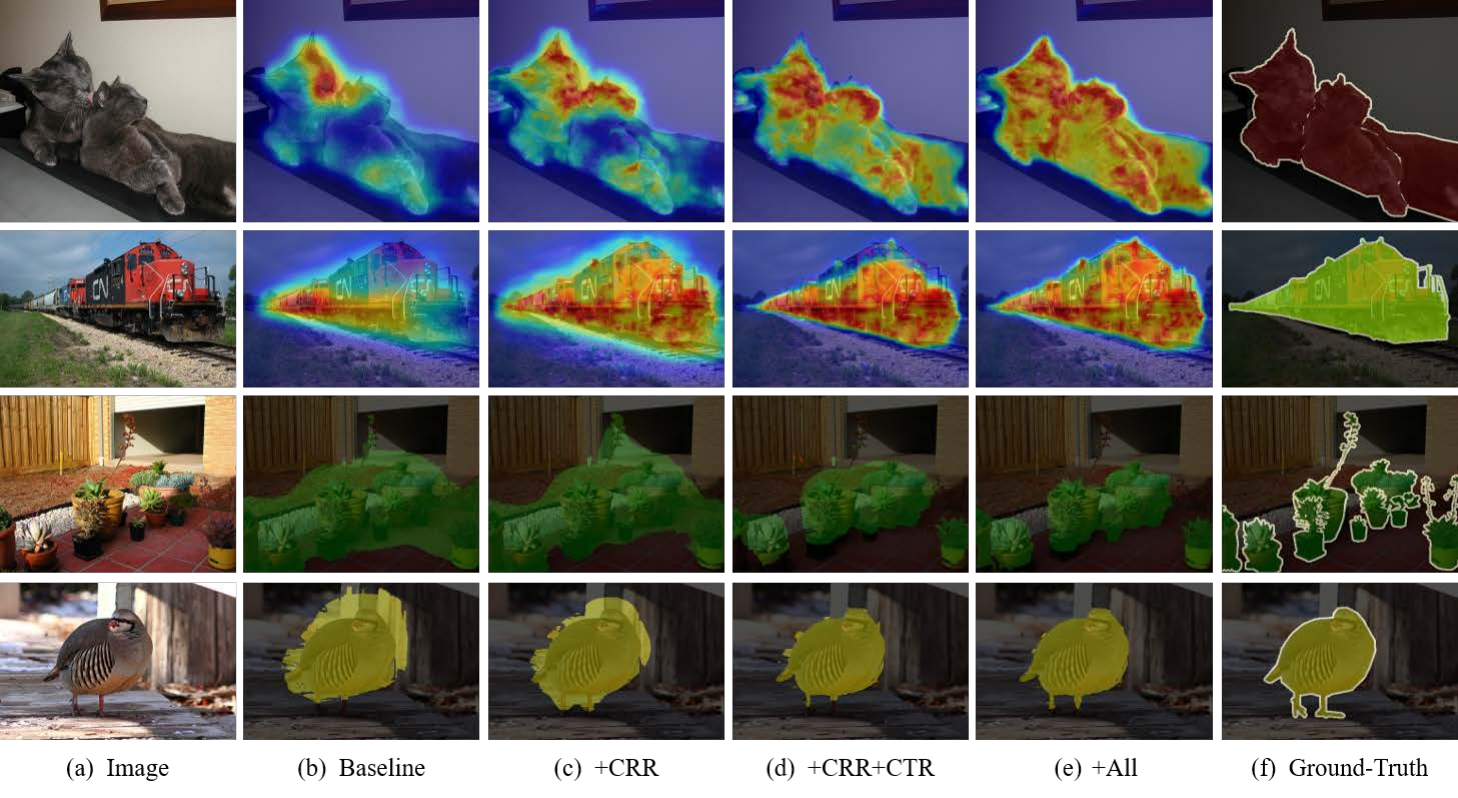}
\vspace{-4mm}
\caption{Visualization examples for results of CAMs and segmentation masks generated by various methods, including the baseline, CRR, CRR + CTR, all (\ie, + CRR + CTR + ComLoss), and ground-truth. These methods were applied to the \emph{train} and \emph{val} sets of PASCAL VOC 2012~\cite{everingham2010pascal}. The results demonstrate the effectiveness of our proposed modules in progressively addressing under-activation or over-activation issues and identifying low-confident object regions.}
\label{fig4}
\vspace{-4mm}
\end{figure*}

%% file: figs_tables/table1.tex
\begin{table*}[t] 
\begin{center}
\renewcommand\arraystretch{1.4}
\setlength{\tabcolsep}{3pt}{
\caption{The results of ablation studies conducted to evaluate the quality of class activation maps (CAMs) and segmentation performance on the PASCAL VOC 2012 dataset~\cite{everingham2010pascal}. The input image sizes used for the experiments are fixed at $448\times 448$ and $128\times 128$. The results of \emph{train (CAMs)} and \emph{val (CAMs)} refer to the CAMs produced by the classification branch, and false positive classes are removed using the ground-truth of image-level labels. The final segmentation performance is evaluated using the results of \emph{val} masks without providing the ground-truth of image-level labels. The efficiency of our proposed method is demonstrated through the parameter count (Params) and computational complexity (FLOPs).}
\begin{tabular}{ c  c  c c | c c | c  c  c } 
Baseline  & CRR & CTR & ComLoss & Params (M) & FLOPs (G) & CAMs(\emph{training})~(\%) & CAMs(\emph{val})~(\%) & Mask(\emph{val})~(\%) \\ 
\hline \hline
\cmark &  &   & &135.94 & 498.24 & 50.10  & 48.69  & 47.65  \\
\cmark & \cmark&  &  &136.67$_{\color{red}{+0.73}}$ & 500.51$_{\color{red}{+2.27}}$ &54.39$_{\color{red}{+4.29}}$ & 52.02$_{\color{red}{+3.33}}$ &52.71$_{\color{red}{+5.06}}$ \\ 
\cmark & \cmark& \cmark &  &136.67$_{\color{red}{+0.73}}$ & 500.51$_{\color{red}{+2.27}}$ & 63.71$_{\color{red}{+13.61}}$ & 62.15$_{\color{red}{+13.46}}$  &57.09$_{\color{red}{+9.44}}$ \\
\cmark & \cmark   & \cmark & \cmark  &136.67$_{\color{red}{+0.73}}$ & 500.51$_{\color{red}{+2.27}}$ &65.50$_{\color{red}{+15.40}}$ & 63.06$_{\color{red}{+14.37}}$ &65.34$_{\color{red}{+17.69}}$ \\ 
\end{tabular}\label{tab1}}
\vspace{-4mm}
\end{center}
\end{table*}

%% file: figs_tables/table2.tex
\begin{table}[t] 
\begin{center}
\renewcommand\arraystretch{1.4}
\setlength{\tabcolsep}{.5pt}{
\caption{The results of ablation studies conducted to evaluate the quality of CAMs and segmentation performance on the validation set of PASCAL VOC 2012~\cite{everingham2010pascal}. ``+'' denotes the deployment of the corresponding scheme on the baseline model. The refinement module used in the SEAM~\cite{wang2020self} method to refine CAMs is denoted as PCM. The term ``STR'' represents Self-Transform Regularization, which adopts the self-view projected features rather than the other-view ones.} 
\begin{tabular}{ l c  c  c | c c c } 
Methods &   &   &   & CAMs(\emph{training}) & CAMs(\emph{val}) & Mask(\emph{val}) \\ 
\hline \hline
baseline &    &    &    & 50.10 & 48.69   & 47.65   \\
\hline\multicolumn{7}{l}{(\textbf{a}) \textbf{\emph{Superiority of CRR}}} \\
+ PCM~\cite{wang2020self} + CTR &    &    &  &59.36$_{\color{red}{+9.26}}$ & 57.19$_{\color{red}{+8.50}}$ &52.19$_{\color{red}{+4.54}}$\\ 
+ CRR + CTR  &    &    &  & 63.71$_{\color{red}{+13.61}}$ & 62.15$_{\color{red}{+13.46}}$  &57.09$_{\color{red}{+9.44}}$ \\
 \hline
\multicolumn{7}{l}{(\textbf{b}) \textbf{\emph{Superiority of CTR}}} \\ 
+ STR + CRR + ComLoss &    &    &   & 63.61$_{\color{red}{+13.51}}$ & 61.74$_{\color{red}{+13.05}}$   &60.38$_{\color{red}{+12.73}}$ \\
+ CTR + CRR + ComLoss &    &    &   &65.50$_{\color{red}{+15.40}}$ & 63.06$_{\color{red}{+14.37}}$   &65.34$_{\color{red}{+17.69}}$ \\
\end{tabular}\label{tab2}}
\vspace{-6mm}
\end{center}
\end{table}

%% file: figs_tables/table3.tex
\begin{table}[t!]
\begin{center}
\renewcommand\arraystretch{1.4}
\setlength{\tabcolsep}{6pt}{
\caption{Comparisons with the state-of-the-art methods on the \emph{val} and \emph{test} sets of Pascal VOC 2012~\cite{everingham2010pascal} in terms of mIoU ($\%$). ``Superv." is the training supervision ($\mathcal{F}$: pixel-level mask, $\mathcal{I}$: image-level class label, $\mathcal{B}$: bounding box label, $\mathcal{P}$: point label, $\mathcal{S}$: scribble label, $\mathcal{SAL}$: salient mask, $\mathcal{D}$: extra data). ``Seg. Backbone'' denotes the backbone network used for the semantic segmentation task. }
\begin{tabular}{ l c c | c c } 
Methods & Seg. Backbone & Superv. & \emph{val}~(\%) & \emph{test}~(\%) \\
\hline \hline
\multicolumn{4}{l}{(\textbf{a}) \textbf{\emph{Large Vision Models}}} \\ 
\cdashline{1-5}[0.8pt/2pt]
SAM~\cite{kirillov2023segany,jiang2023segment} & ResNet-101 & $\mathcal{I}$ & 71.1 & 72.2\\
SAM~\cite{kirillov2023segany,jiang2023segment} & ResNet-101 & $\mathcal{P}$ & 69.0 & 68.7\\
SAM~\cite{kirillov2023segany,jiang2023segment} & ResNet-101 & $\mathcal{S}$ & 75.9 & 76.6\\
SAM~\cite{kirillov2023segany,jiang2023segment} & ResNet-101 & $\mathcal{B}$ & 76.3 &75.8 \\
\hline \hline
\multicolumn{4}{l}{(\textbf{b}) \textbf{\emph{Multi-stage}}} \\ 
\cdashline{1-5}[0.8pt/2pt]
EPS~\cite{lee2021railroad} & VGG-16 &$\mathcal{I,SAL}$  & 66.60  & 67.90 \\
OAA+~\cite{jiang2021online} &  ResNet-101 & $\mathcal{I,SAL}$ & 66.10  & 67.20\\
IRN~\cite{ahn2019weakly} & ResNet-50 & $\mathcal{I}$  & 63.50   & 64.80 \\
SEAM~\cite{wang2020self} & WideResNet-38 &  $\mathcal{I}$  & 64.50  & 65.70  \\
CONTA~\cite{zhang2020causal} & ResNet-50 & $\mathcal{I}$  &  65.30   & 66.10 \\
ReCAM~\cite{chen2022class} & ResNet-101 &  $\mathcal{I}$  & 68.40  & 68.20  \\
 MCIS~\cite{wang2022looking} &  ResNet-101 &  $\mathcal{I}$  &  66.20  & 66.90  \\
RIB~\cite{lee2021reducing} & ResNet-101 & $\mathcal{I}$  &  68.30 & 68.60\\
MCTformer~\cite{xu2022multi} & WideResNet-38 & $\mathcal{I}$  &  71.90 & 71.60 \\
SIPE~\cite{chen2022self} & WideResNet-38 & $\mathcal{I}$  &  68.20 & 69.50\\
W-OoD~\cite{lee2022weakly}& WideResNet-38 & $\mathcal{I}$  &  70.70 & 70.10\\
ESOL~\cite{li2022expansion} & ResNet-101 & $\mathcal{I}$  &  69.90 & 69.30\\
PPC~\cite{du2022weakly}& WideResNet-38 & $\mathcal{I}$  &  67.70 & 67.40 \\
ACR~\cite{kweon2023weakly}& WideResNet-38 & $\mathcal{I}$  &  71.90 & 71.90 \\
\cdashline{1-5}[0.8pt/2pt]
\textbf{RTC} w/o CRF & WideResNet-38 & $\mathcal{I}$  & \textbf{69.45}   & \textbf{70.01}  \\
\textbf{RTC} & WideResNet-38 & $\mathcal{I}$  & \textbf{70.21}   & \textbf{70.76} \\
\textbf{RTC} w/o CRF & ResNet-101 & $\mathcal{I}$ &\textbf{71.08} & \textbf{72.02} \\
\textbf{RTC} & ResNet-101  & $\mathcal{I}$  &\textbf{71.56} & \textbf{72.33} \\
\hline \hline
\multicolumn{4}{l}{(\textbf{c}) \textbf{\emph{Single-stage}}} \\ 
\cdashline{1-5}[0.8pt/2pt]
EM~\cite{papandreou2015weakly} & VGG-16 & $\mathcal{I}$  & 38.20  & 39.60  \\
CRF-RNN~\cite{roy2017combining} & VGG-16 & $\mathcal{I}$  & 52.80   & 53.70 \\
RRM~\cite{zhang2020reliability}& WideResNet-38 & $\mathcal{I}$  & 62.60  & 62.90 \\
SSSS~\cite{araslanov2020single} & WideResNet-38 & $\mathcal{I}$  & 62.70  & 64.30 \\
AA\&LR~\cite{zhang2021adaptive} & WideResNet-38 & $\mathcal{I}$  & 63.90  & 64.80 \\
AFA~\cite{ru2022learning} & MiT-B1 & $\mathcal{I}$  & 66.00  & 66.30 \\
SLRNet~\cite{pan2022learning} & WideResNet-38 & $\mathcal{I}$  & 67.20  & 67.60 \\
TSCD~\cite{xu2023self}&  MiT-B1  & $\mathcal{I}$  & 67.30  & 67.50 \\
ToCo~\cite{ru2023token} &  MiT-B1  & $\mathcal{I}$  & 71.10  & 72.20 \\
PPC$^\S$~\cite{du2022weakly} & WideResNet-38 & $\mathcal{I}$  & 58.18  & 59.67 \\
\cdashline{1-5}[0.8pt/2pt]
\textbf{RTC} w/o CRF & WideResNet-38 & $\mathcal{I}$  & \textbf{65.34}   & \textbf{66.90}  \\
\textbf{RTC} & WideResNet-38 & $\mathcal{I}$  & \textbf{67.20}   & \textbf{68.76} \\
\hline \hline
\end{tabular}
\label{tab3}}
\vspace{-6mm}
\end{center}
\end{table}

%% file: figs_tables/fig5.tex
\begin{figure*}[tb]
\includegraphics[width=.98 \textwidth]{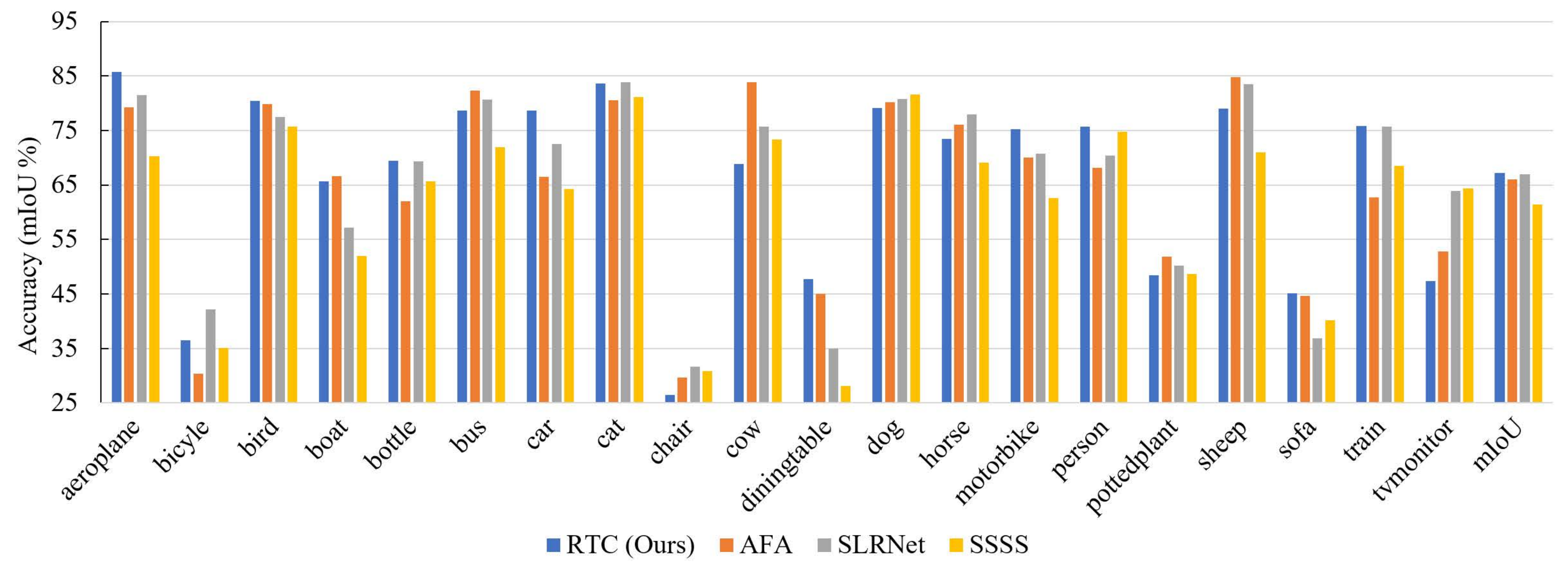}
\vspace{-4mm}
\caption{Visualization examples of segmentation results in terms of class IoU generated by SSSS~\cite{araslanov2020single}, AFA~\cite{ru2022learning}, SLRNet~\cite{pan2022learning} and our method on the \emph{val} set of PASCAL VOC 2012~\cite{everingham2010pascal}. Compared to the other methods, our method can learn more complete and accurate object regions on most classes, which demonstrates the effectiveness of our \textbf{RTC} in improving the qualites of CAMs as pseudo-masks online and the segmentation qualities. }
\label{fig5}
\vspace{-4mm}
\end{figure*}

%% file: figs_tables/table4.tex

\begin{table*}[tb]
\begin{center}
\renewcommand\arraystretch{1.4}
\setlength{\tabcolsep}{8pt}{
\caption{Comparsion of MaxBoxAccV2 scores and localization performance on CUB-200-2011~\cite{welinder2010caltech} based on VGG16~\cite{simonyan2015very} and ResNet50~\cite{he2016deep}. ``-" denotes that there is no reported result in its paper.}
\vspace{-2mm}
\begin{tabular}{l|c|cccc|ccc}
\hline
 \multirow{2}{*}{Method} & \multirow{2}{*}{Backbone} & \multicolumn{3}{c}{$\delta$} & \multirow{2}{*}{Mean$~(\%)$} & \multirow{2}{*}{Loc-1$~(\%)$ } & \multirow{2}{*}{Loc-5$~(\%)$} & \multirow{2}{*}{GT Loc$~(\%)$} \\
 &  & 0.3$~(\%)$ & 0.5$~(\%)$ & 0.7$~(\%)$ &  &  &  &  \\ \hline
CutMix~\cite{yun2019cutmix} & VGG16 & 91.10 & 67.30 & 28.60 & 62.30 & 43.45 & - & - \\
Ki et al~\cite{ki2020sample} & VGG16 & 96.20 & 77.20 & 26.80 & 66.70 & 57.50 & - & - \\
CREAM~\cite{xu2022cream} & VGG16 & - & - & - & 62.20 & 70.44 & 85.67 & 90.98 \\
KD-CI-CAM~\cite{shao2023further} & VGG16 & - & - & - & - & 73.00 & 86.50 & 91.60 \\
PDM~\cite{meng2022diverse}& VGG16 & - & - & - & - & 67.30 & 82.20 & 82.20 \\
Counterfactual-CAM~\cite{shao2023mitigating} & VGG16 & - & - & - & - & \textbf{73.70} & - & 91.60\\
Bridging~\cite{kim2022bridging} & VGG16 & 99.30 & 93.20 & 47.80 & 80.10 & 70.83 & 88.07 & 93.17 \\
\hline
\textbf{Bridging+Ours} & VGG16 & \textbf{99.36} & \textbf{93.84} & \textbf{57.00} & \textbf{84.31} & 70.76 & \textbf{88.25} & \textbf{93.91} \\
\hline \hline
CutMix~\cite{yun2019cutmix} & ResNet50 & 94.30 & 71.50 & 22.50 & 62.80 & 54.81 & - & - \\
Ki et al~\cite{ki2020sample} &  ResNet50 & 96.20 & 72.80 & 20.60 & 63.20 & 56.10 & - & - \\
CREAM~\cite{xu2022cream} & ResNet50  &-  &-  & - &64.90  & \textbf{76.03}  & - & 89.88 \\
BagsCAM~\cite{zhu2022bagging}& ResNet50 &-  &-  & - &\textbf{84.88} & 69.67  & -  &  \textbf{94.01}\\
PDM~\cite{meng2022diverse}& ResNet50  & - & - & - & - & 71.20 & 83.60 & 82.30 \\
Bridging~\cite{kim2022bridging} & ResNet50 & 99.40 & 90.40 & 38.00 & 75.90 & 73.16 & 86.68 & 91.60 \\
\hline
\textbf{Bridging+Ours} & ResNet50 & \textbf{99.40} & \textbf{92.20} & \textbf{54.00} & 81.90 & 74.01 & \textbf{87.87} & 92.32 \\
\hline
\end{tabular}
\label{tab4}}
\vspace{-5mm}
\end{center}
\end{table*}

%% file: figs_tables/fig6.tex
\begin{figure}[tb]
\includegraphics[width=.48 \textwidth]{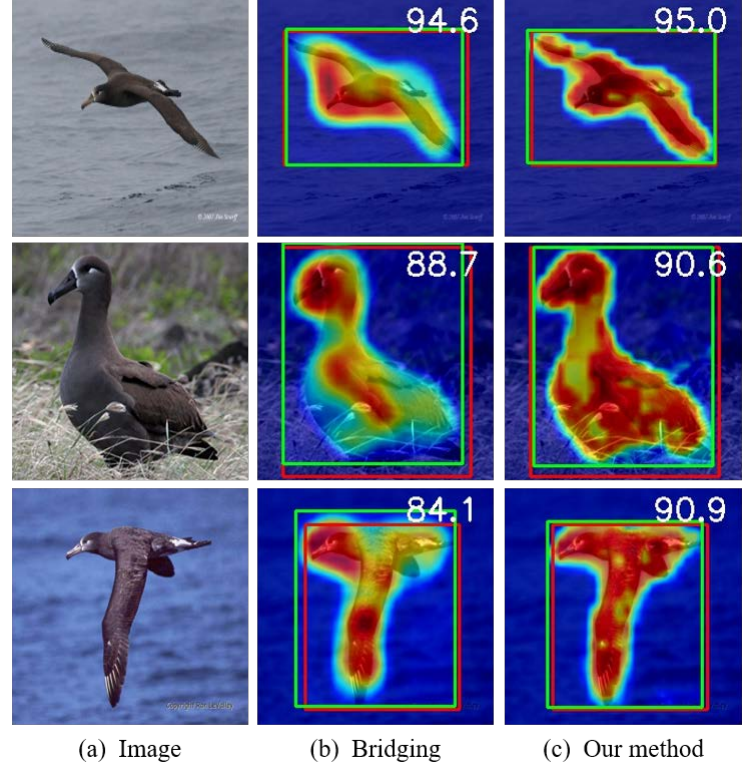}
\vspace{-4mm}
\caption{Visualizations of location performance on CUB-200-2011~\cite{welinder2010caltech} based on the ResNet50. Green and red boxes represent the predicted bounding boxes and ground-truth, respectively. The number means the IoU value of Loc-1. These visualization results demonstrate our proposed method can discover the low-confident object regions and locate the object more accurately.}
\label{fig6}
\vspace{-4mm}
\end{figure}

%% file: sections/5_conclusion.tex
\section{Conclusion}
In this paper, we proposed a novel approach to refine CAMs by utilizing different feature representations from the backbone and employing cross-transformed features to learn robust class representations. This approach can help identify low-confident object regions and alleviate the problems of under-activation and over-activation. Additionally, the compensatory loss improves the CAMs to provide high-quality and stable pseudo-masks for supervising the segmentation result. The effectiveness of our method was validated through extensive experiments. In future research, we plan to explore adaptive selection of the quantity-quality trade-off pixels to construct class prototypes, which can further improve the accuracy of contrastive learning and produce precise class representations with strong generalization ability. Additionally, we will investigate the use of a transformer architecture, which can capture more global context information than the CNN architecture, to improve feature representation and learn high-quality pseudo-masks in WSSS.